\documentclass{article}
\usepackage{amssymb}
\usepackage{multicol}
\usepackage[bookmarks=true]{hyperref}
\usepackage{amsmath}
\usepackage{amsfonts}
\usepackage{graphicx}
\usepackage{xcolor}
\usepackage{booktabs}
\usepackage{wrapfig}
\usepackage[font=scriptsize]{caption}
\usepackage{subcaption}
\usepackage{amsthm}
\usepackage{scontents}

\theoremstyle{definition}

\usepackage[style=ieee, citestyle=numeric-comp, backend=biber]{biblatex}

\usepackage[preprint, nonatbib]{corl_2025}

\makeatletter
\newif\ifcorlfinal
\@ifclasswith{corl_2025}{final}{\corlfinaltrue}{\corlfinalfalse}
\@ifpackagewith{corl_2025}{final}{\corlfinaltrue}{}
\makeatother

\title{Mobi-$\pi$: Mobilizing Your Robot Learning Policy}

\author{
    Jingyun Yang$^{\dagger1}$, Isabella Huang\thanks{~These authors contributed equally. \ $^\dagger$Contact: \texttt{jingyuny@stanford.edu}}$^{*2}$, Brandon Vu$^{*1}$\\
    \textbf{Max Bajracharya$^2$, Rika Antonova$^3$, Jeannette Bohg$^1$} \\
    $^1$Stanford University \qquad $^2$Toyota 
    Research Institute \qquad $^3$University of Cambridge
}

\newcommand{\name}{Mobi-$\pi$}

\definecolor{my_red}{HTML}{FF555D}
\definecolor{my_green}{HTML}{1EAE94}
\definecolor{my_blue}{HTML}{3598DA}
\definecolor{my_purple}{HTML}{B298D2}
\definecolor{my_orange}{HTML}{FDA553}
\definecolor{linkcolor}{HTML}{3598DA}
\definecolor{citecolor}{HTML}{1EAE94}
\definecolor{my_gray}{HTML}{7f8c8d}

\hypersetup{
    colorlinks = true,
    linkcolor = linkcolor,
    urlcolor = linkcolor,
    citecolor = citecolor
}

\begin{document}
\maketitle

\begin{abstract}
Learned visuomotor policies are capable of performing increasingly complex manipulation tasks. 
However, most of these policies are trained on data collected from limited robot positions and camera viewpoints.
This leads to poor generalization to novel robot positions, which limits the use of these policies on mobile platforms, especially for precise tasks like pressing buttons or turning faucets.
In this work, we formulate the {\em policy mobilization\/} problem: find a mobile robot base pose in a novel environment that is in distribution with respect to a manipulation policy trained on a limited set of camera viewpoints. 
Compared to retraining the policy itself to be more robust to unseen robot base pose initializations, policy mobilization decouples navigation from manipulation and thus does not require additional demonstrations. 
Crucially, this problem formulation complements existing efforts to improve manipulation policy robustness to novel viewpoints and remains compatible with them. 
We propose a novel approach for policy mobilization that bridges navigation and manipulation by optimizing the robot's base pose to align with an in-distribution base pose for a learned policy. Our approach utilizes 3D Gaussian Splatting for novel view synthesis, a score function to evaluate pose suitability, and sampling-based optimization to identify optimal robot poses. To understand policy mobilization in more depth, we also introduce the \name{}~framework, which includes: (1) metrics that quantify the difficulty of mobilizing a given policy, (2) a suite of simulated mobile manipulation tasks based on RoboCasa to evaluate policy mobilization, and (3) visualization tools for analysis. In both our developed simulation task suite and the real world, we show that our approach outperforms baselines, demonstrating its effectiveness for policy mobilization.
Website: \href{https://mobipi.github.io}{\texttt{https://mobipi.github.io}}
\end{abstract}
\begin{figure}[h!]
    \centering
    \vspace{5pt}
    \includegraphics[width=\linewidth]{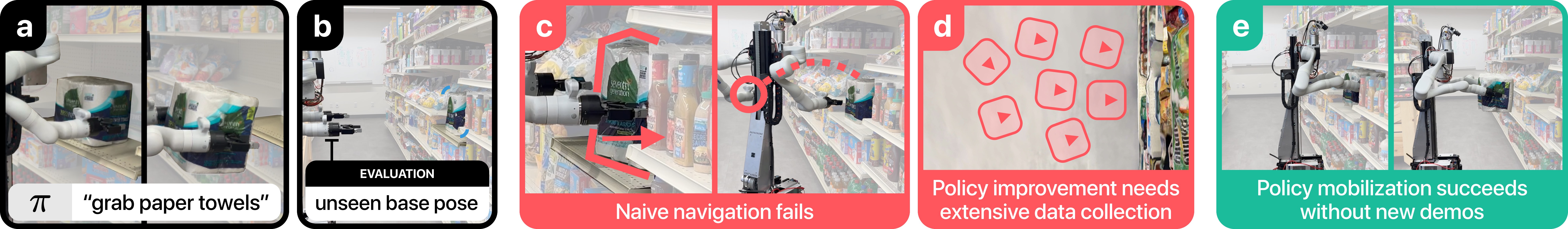}
    \caption{\textbf{Introducing policy mobilization.} (a) Assume a visuomotor policy $\pi$ trained from one or a set of limited camera poses. (b) We are interested in running $\pi$ on a mobile platform, where the robot is initialized at a random pose and needs to navigate before running $\pi$. (c) \textbf{\color{my_red}Naively navigating towards the object of interest} and executing the manipulation policy is likely to result in an out-of-distribution initialization for the policy, leading to failures. Left: The robot gets too close and pushes the object inward with the arm. Right: suboptimal heading makes the object unreachable for the left arm. (d) \textbf{\color{my_red}Improving the robustness of the policy} requires extensive data collection to cover all possible robot base pose initializations. (e) We define the new problem of \textbf{\color{my_green}policy mobilization} which aims to find the optimal robot pose that leads to an in-distribution viewpoint for executing $\pi$, achieving task success without the need to collect additional demonstrations.}
    \vspace{-7pt}
    \label{fig:nav_does_not_work}
\end{figure}

\section{Introduction}

Recent advances in learning visuomotor manipulation policies have enabled robots to execute increasingly complex manipulation tasks using imitation and reinforcement learning~\cite{chi2023diffusionpolicy,Zhao-RSS-23,brohan2022rt,brohan2023rt,team2024octo,kim2024openvla,black2024pi_0}. However, most manipulation policies are trained on data collected from a limited set of robot base poses and camera viewpoints, leading to a tight coupling between the policy's input distribution and the configurations seen during training. When such a policy is deployed on a mobile robot, changes in the robot's base pose inherently alter its viewpoint. This can cause both the visual input and the reachable workspace to drift outside the policy's training distribution, severely degrading performance. 

Given the abundance of stationary robot data, it is highly desirable to reuse existing datasets and pre-trained policies for mobile manipulation. While retraining a policy to handle a wide range of base poses and viewpoints is a straightforward solution, it is also costly—requiring extensive data collection across diverse positions and scene variations~\cite{brohan2022rt, khazatsky2024droid}. Moreover, as policies are deployed in mobile settings, their field of view expands, demanding stronger robustness to scene-level variations.
An alternative strategy is to decouple navigation and manipulation by chaining a navigation module with the trained manipulation policy.
However, most prior work in navigation specifies goals via language instructions or object locations~\cite{zhang2018semantic, rosinol2020kimera, chang2020semantic, ye2021auxiliary, deitke2022, gireesh2022object, gadre2022clip, ramrakhya2023pirlnav, hiroselelan, yokoyama2024vlfm} rather than selecting base poses that are {\em policy-compatible\/}. As a result, the robot often navigates to viewpoints that lie outside of the policy's training distribution, leading to unreliable performance.

In this work, we introduce the \emph{policy mobilization problem\/}: determine a mobile robot base pose in a novel environment that aligns with the training distribution of a manipulation policy learned from a limited set of camera viewpoints. Addressing this problem decouples navigation from manipulation policy training while ensuring that navigation remains {\em policy-aware\/}—that is, it selects poses that are compatible with the policy’s visual input distribution. This enables effective policy deployment in unseen environments without requiring extensive additional data collection. Consequently, policy mobilization methods enhance the viewpoint robustness of manipulation policies, independent of the policy's original tolerance to viewpoint variations.

As a first attempt to tackle the policy mobilization problem, we propose a method that searches for robot base poses likely to yield successful policy execution. Our approach (a) represents the scene using 3D Gaussian Splatting~\cite{kerbl3Dgaussians}, (b) uses differentiable rendering to evaluate whether a candidate pose is in-distribution, provides visibility of task-relevant objects, and avoids collisions, and (c) optimizes over poses with a sampling-based procedure to identify the most suitable robot base location.
To systematically study the policy mobilization problem, we present the \textit{\name{} framework}, which includes (1) metrics to quantify the difficulty of mobilizing a given policy, (2) a suite of simulated mobile manipulation tasks built on RoboCasa~\cite{robocasa2024}, (3) visualization tools for qualitative analysis.

We first evaluate {\em non-policy aware\/} baselines that decouple navigation and manipulation without considering that policies perform better from specific base poses. These approaches frequently fail due to the poor alignment with the policy's training distribution. We also examine {\em policy-aware\/} baselines that learn to connect navigation and manipulation using large amounts of additional training data and show that they still struggle to generalize to unseen room layouts.
Finally, we validate our approach on both the \name{} simulation suite and three real-world mobile manipulation tasks. Our method outperforms all baselines in various simulation and real-world experiments, enabling effective deployment of manipulation policies trained solely on stationary robot data.
\section{Related Work}

\textbf{Semantic navigation.}
In semantic navigation, robots navigate environments to find specific objects or regions. Prior work~\cite{kuipers2000spatial, kuipers2004local, aydemir2013active, zhang2018semantic, zheng2019pixels, rosinol2020kimera, jin2023focusing} uses classical methods like SLAM~\cite{kazerouni2022survey} to build and explore semantic maps, while recent methods use learning-based approaches~\cite{chang2020semantic, ye2021auxiliary, deitke2022, gireesh2022object,ramrakhya2023pirlnav, hiroselelan}. Other methods predict semantic top-down maps~\cite{ramakrishnan2022poni, chen2023not, yokoyama2024vlfm} and navigate using waypoint planners~\cite{hughes2022hydra}. These approaches typically stop in an area where the object is visible and close, without ensuring that the viewpoint is suitable for manipulation. In contrast, our method selects navigation targets that maximize downstream manipulation policy success.

\textbf{Imitation learning.}
Imitation learning teaches robots tasks from demonstrations~\cite{levine2016end, zeng2021transporter, shafiullah2022behavior, chi2023diffusionpolicy, aldaco2024aloha, fu2024mobile, chi2024universal}, with recent efforts scaling to generalist models~\cite{brohan2022rt, brohan2023rt, gu2023rt, yang2024equivact, team2024octo, khazatsky2024droid, kim2024openvla, yang2024equibot, black2024pi_0}. Training such policies for both navigation and manipulation requires massive data. Instead, we use imitation learning for the manipulation policy and leverage existing navigation methods to move the robot to an in-distribution robot pose ensuring policy success. Our insight is that by selecting suitable poses for manipulation, we reduce the need for collecting more data from many robot poses or viewpoints and improve deployment in novel scenes. We focus on single-task policies and leave multi-task extensions to future work.

\textbf{Mobile manipulation.}
Prior work has studied various ways to bring learned policies to mobile manipulation~\cite{saycan2022arxiv, wu2023tidybot, shafiullah2023bringing, xiong2024adaptive, fu2024mobile, qiu2024wildlma, wu2024tidybot++, huang2024points2plans}. Some methods~\cite{brohan2022rt,fu2024mobile} rely on end-to-end learning but require large-scale data. Others~\cite{saycan2022arxiv, wu2023tidybot, shafiullah2023bringing, qiu2024wildlma} assume known initial base poses for a given task. ARPlace~\cite{2012_Freek} predicts valid robot poses for running skills but focuses more on reachability rather than viewpoint restrictions for these skills. ASC~\cite{yokoyama2023asc} trains coordinated mobile pick-and-place skills with RL in simulation but requires highly accurate simulators, which are often difficult to obtain for complex real-world tasks with deformables or complex multi-object interactions. Some works use reachability maps or abstract planners~\cite{jauhri2022robot, rosen2023synthesizing}, but ignore that policies may be viewpoint sensitive. 
Our approach directly addresses this by finding robot base and camera poses that enable manipulation policy success. This ``mobilizes'' policies with fixed initial base poses, making them usable on mobile robots in complex environments, including those with deformable objects or multi-object interactions.

\textbf{In-distribution detection.}
Policy mobilization involves finding an in-distribution initial condition for a given policy. Prior work has extensively studied in-distribution detection or anomaly detection, covering classification-based methods~\cite{hendrycks2016baseline, liang2017enhancing, ruff2018deep, golan2018deep, fort2021exploring}, reconstruction methods~\cite{hendrycks2018deep, nalisnick2018deep}, methods that utilize generative adversarial networks (GANs)~\cite{di2019survey, schlegl2019f}, and diffusion-based methods~\cite{liu2025survey}. Nearly all these methods assume a passive setting where they detect whether a given input is in-distribution without solving the problem of how an agent can find a sample that is more in-distribution.
Recent robotics works have started to address out-of-distribution (OOD) detection~\cite{farid2022task, sinha2022system, agia2024unpacking} but treat the perception module as a passive component without actively updating the robot's camera or base pose to mitigate OOD scenarios. Our work closes this gap, marrying in-distribution detection with active robot pose optimization in a mobile manipulation context.

\textbf{Hierarchical methods for policy learning.}
Our setting resembles Hierarchical Reinforcement Learning (HRL), where an option~\cite{sutton1999between} includes an initiation set, a policy, and a termination function. HRL methods~\cite{bacon2017option, machado2017laplacian, frans2017meta, bagaria2021robustly, bagaria2023effectively} often learn these jointly but have limited success in complex settings. Instead of learning all elements of an option, we focus on selecting optimal robot poses for pre-trained manipulation policies in realistic environments.

\textbf{Training view-robust policies.}
Related to the problem setting of our work, prior work has studied methods of obtaining view-robust manipulation policies. One approach achieves this by collecting multi-view data in simulation~\cite{sadeghi2018sim2real, hirose2023exaug, seo2023multi}, but suffers from the sim-to-real transfer challenge. Some works achieve view-robustness by using 3D representations, but training a policy to use 3D input requires data with well-calibrated camera extrinsics~\cite{gervet2023act3d, zhu2023learning, ze20243d}, which is not readily available in many robot learning datasets today. Some works achieve real-world view robustness by manually collecting multi-view data in the real world~\cite{khazatsky2024droid}, but this requires extensive human effort. There are proposals for using single-view novel view synthesis to bypass the need for multi-view data~\cite{tian2024viewinvariant}, but methods for novel view synthesis from a single image struggle to effectively render views that are too different from the given viewpoint. From a view robustness perspective, our work takes advantage of the mobility of the robot so the robot can navigate to a base pose with an appropriate camera viewpoint for subsequent policy execution, thus requiring no extensive additional data collection.
\section{The Policy Mobilization Problem}

Let the environment be an MDP $(\mathcal{S}, \mathcal{A}, \mathcal{P}, \mathcal{R})$. $\mathcal{S}$ is the state space, where a state $s = (s^\text{env}, p, q)$ consists of the environment state $s^\text{env}$, the mobile base pose $p \in SE(2)$, and the arm configuration $q \in \mathbb{R}^{n_\text{dof}}$ with $n_\text{dof}$ denoting the number of degrees of freedom of the robot arms. $\mathcal{A}$ is an action space that consists of base and arm actions. $\mathcal{P}: \mathcal{S} \times \mathcal{A} \times \mathcal{S} \rightarrow [0, +\infty)$ represents the state transition probability. $\mathcal{R}: \mathcal{S} \rightarrow \mathbb{R}$ is the reward function. 
We are given a pre-trained manipulation policy $\pi: \mathcal{O} \rightarrow \mathcal{A}$, where $\mathcal{O}$ is the observation space.
We measure performance of the mobilized policy with the expected final reward $J(\pi|s_\text{start}) = \mathbb{E}_{\tau \sim \pi}\big[ \mathcal{R}(s_H)|s_0 = s_\text{start} \big]$, where $\tau = (s_0, a_0, s_1, \cdots, s_H)$ is a trajectory of length $H$ generated by executing $\pi$ from state $s_\text{start}$. 
The initial state is $s_0 = (s^{\text{env}}_0, p_0, q_0)$. Policy mobilization aims to find an optimal base pose $p^*$ to run the manipulation policy such that the expected return is maximized:
$p^* = \text{argmax}_{p \in \text{SE}(2)} J\big(\pi | (s^{\text{env}}_0, p, q_0)\big).$

\section{A Novel Method for Policy Mobilization}
\label{sec:policy_mobilization}

\begin{figure*}
    \centering
    \includegraphics[width=\linewidth]{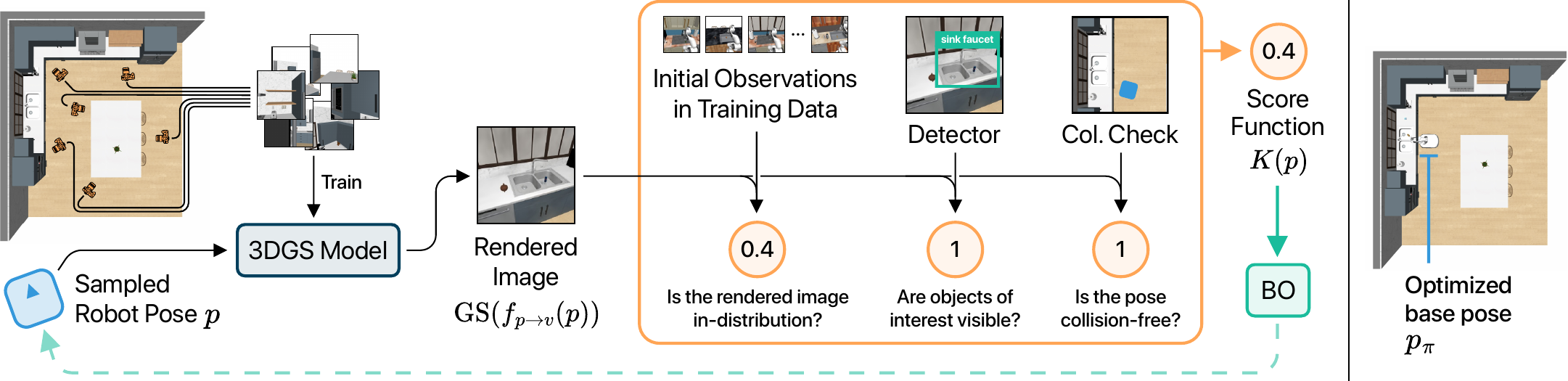}
    \vspace{-13pt}
    \caption{\textbf{Overview of our proposed proof-of-concept method.}
    The goal of our method is to find a proper robot pose $p$ for policy initialization such that a given policy $\pi$ can be successfully executed in a scene $S$.
    Our method has three components:
    (1) a proposed robot pose $p$ is converted to its corresponding camera pose $v$ via $v = f_{p \to v}(p)$ and then sent into a 3D Gaussian Splatting model to produce a rendered image;
    (2) the rendered image and the proposed robot pose are passed into a hybrid score function $K$ that predicts whether the proposed robot pose is suitable for policy initialization;
    (3) the score function output is used to update a Bayesian Optimization algorithm, which repeatedly proposes new sampled robot poses to be evaluated and updates its internal belief on which robot poses are optimal for robot policy execution.
    The hybrid score function is composed of multiple parts powered by visual foundation models so that it can make judgments of whether the proposed pose (a) is in distribution with respect to the training dataset;
    (b) has the objects of interest in view; and (c) is collision-free. 
    }
    \vspace{-15pt}
    \label{fig:method_overview}
\end{figure*}

We introduce a novel method for policy mobilization, illustrated in Figure~\ref{fig:method_overview}.
We want our method to satisfy the following criteria: (1) take into account the capabilities of the policy, (2) do not require extensive data collection for policy learning, and (3) operate in arbitrary unseen room layouts.

\textbf{Assumptions.} We need a setting where our method is aware of the policy's capabilities without collecting large-scale navigation data. To make this possible, we assume access to the initial image frames of each demonstration in the training dataset: $\mathcal{D}_{\text{initial}} = \{[o_1^1, \ldots, o_1^{N_{DS}}]\}$ used for training the policy,
where $o_t^k$ is the $t$-th observation of the $k$-th training episode and
$N_{DS}$ denotes the number of trajectories in the dataset.
To successfully operate in any unseen room layout, our method needs a mechanism for quickly learning about the test-time room layout. Therefore, we assume that before executing any policy evaluation, our method is allowed to build a map of the scene by looking around the scene \textit{without} interacting with it.
For simplicity, we assume that the initial robot arm pose is fixed and known for each task. We also assume known camera intrinsics and a known, fixed relative transform between the robot base and the camera. This implies that optimizing robot base poses is the same as optimizing camera viewpoints in our setting.
At the start of a mobile manipulation episode, the mobile robot is placed at a pose with an unobstructed view of the target object $p_{\text{init}} \in SO(2)$.
The goal is to navigate to a target pose $p_{\pi}$ where the policy $\pi$ can be successfully executed.

\textbf{Representing the scene with 3D Gaussian Splatting.} 
When the robot enters a previously unseen scene, it must first acquire a representation of the environment before identifying base poses from which the manipulation policy $\pi$ can be executed effectively.
To this end, we employ 3D Gaussian Splatting (3DGS)~\cite{kerbl3Dgaussians}, a technique that enables high-quality, novel-view synthesis from arbitrary viewpoints.
Given a set of RGB-D images captured from various viewpoints, 3DGS constructs a scene representation composed of Gaussians $\mathcal{G} = \{g_i\}_{i=1}^{N_g}$ where each Gaussian $g_i = (\mu_i, \Sigma_i, o_i, c_i)$ encodes the 3D position $\mu_i \in \mathbb{R}^3$, spatial uncertainty via covariance matrix $\Sigma_i \in \mathbb{R}^{3 \times 3}$, opacity $o_i \in \mathbb{R}$, and view-dependent color $c_i \in SH(4)$.
To render an image from a novel camera pose $v \in SO(3)$, each Gaussian is projected to the image plane~\cite{zwicker2001ewa} and splatted using $\alpha$-blending to generate the final image $I = \text{GS}(v)$. 

We construct the 3DGS model using 1,000 RGB-D images collected throughout the scene. This image acquisition process takes less than 5 minutes and involves driving the robot around the scene while varying the camera height and viewing angle. Future work may consider automating this process so that no human is involved for this image collection. We then reconstruct a point cloud from the depth images relative to the world frame to initialize the Gaussians.
Finally, we train the 3DGS models using \texttt{nerfstudio}~\cite{nerfstudio} for 30,000 update steps.

\textbf{Hybrid score function for estimating policy execution viability.} 
To estimate how likely a policy will be successful from a certain base pose $p$, we define a hybrid score function $K(p)$ that incorporates three key factors:
(1) the rendered view should be in-distribution with respect to the training data,
(2) the object of interest should be visible;
(3) the robot pose should be free of collisions.
To evaluate whether a rendered view is in-distribution, we leverage the initial camera observations $\{o_1^k\}_{k=1}^{N_{DS}}$ in the demonstration dataset. We compare the candidate view against these initial camera observations in a learned feature space. Among the image embeddings we evaluated, dense descriptors from DINO~\cite{caron2021emerging} proved most effective in distinguishing in- from out-of-distribution views and are robust to variations in scene texture.
We precompute DINO features for all initial image observation, yielding$\{\text{DINO}(o_1^k)\}_{k=1}^{N_{DS}}$.
For a queried robot pose $p$, we render the corresponding image using our 3DGS models and compute its dense descriptors as $\text{DINO}(\text{GS}(f_{p \to v}(p)))$, where $f_{p \to v}$ denotes the transformation from base pose $p$ to camera pose $v$. The encoder $\text{DINO}(I)$ maps an image $I$ of size $H \times W$ to a feature tensor in $H' \times W' \times D$. To assess similarity, we apply K-nearest neighbors (KNN) in the feature space to compute the average L2 distance to the demonstration features, yielding an in-distribution score: \begin{equation}{K_{\text{id}}(p) = \text{KNN}\Big(\text{DINO}\big(\text{GS}(f_{p \to v}(p))\big), \{\text{DINO}(o_1^k)\}_{k=1}^{N_{DS}}\Big).}\end{equation}
To increase robustness, we augment $K_{\text{id}} \in \{0, 1\}$ with two additional binary checks.
First, $K_\text{id}$ can return high scores for views that are texturally similar to the training set, even if they fail to include the task relevant object.
To mitigate this, 
we verify object visibility using a vision-language model MiniCPM-v2~\cite{yao2024minicpm} and define  \begin{equation}K_\text{obj}(p) = \text{MiniCPM-v2}(\text{GS}(f_{p \to v}(p))),\end{equation} where $K_\text{obj}(p) = 1$ if the object is detected and $0$ otherwise. We found this approach to outperform conventional object detectors~\cite{minderer2022simple, liu2024grounding}.
Second, we verify that pose $p$ is not in collision with the environment by consulting an occupancy map constructed from the depth data used to train the 3DGS model. We define:
\begin{equation}K_\text{col}(p) = (1\text{  if pose }p\text{ is vacant else }0).\end{equation} 

The final hybrid score combines all three components: \begin{equation}K(p) = 
\begin{cases} 
K_{\text{id}}(p) & \text{if } K_\text{obj}(p) = 1 \text{ }\& \text{ } K_\text{col}(p) = 1, \\
0 & \text{otherwise}.
\end{cases}.\end{equation}

\textbf{Robot pose optimization with Bayesian Optimization.}
Given the score function $K(p)$, our objective is to find the robot base pose $p_{\pi}$ that maximizes the likelihood of successful policy execution: $p_{\pi}= \text{max}_{p}K(p)$.
While some components of $K(p)$ are differentiable, we observe that the overall optimization landscape is non-smooth. In particular, small changes in the camera viewpoint can lead to abrupt variations in the score due to discontinuities introduced by occlusions, changes in object visibility or reachability. These factors make gradient-based optimization unreliable.
To address this, we adopt a gradient-free optimization approach and use Bayesian Optimization (BO) to efficiently search for the optimal base pose $p_{\pi}$.

Bayesian Optimization is a sampling-based, gradient-free optimization algorithm designed to find parameters $\mathbf{\mathbf{w}}^*$ that maximize a target function $f_\text{BO}(\mathbf{w})$. It achieves this by modeling  $f_\text{BO}$ using a surrogate model—typically a Gaussian Process (GP) or random forest—that approximates the underlying function based on observed samples. At each iteration, BO selects new samples from $f_\text{BO}$ by maximizing an acquisition function, such as the Upper Confidence Bound (UCB)~\cite{srinivas2009gaussian}, which balances exploration and exploitation. 
The acquisition function prioritizes regions of the parameter space that are both uncertain and potentially high-scoring, enabling BO to identify promising regions efficiently with a limited number of queries.
To optimize the policy optimization score $K(p)$, we initialize BO with  $N^{(\text{BO})}_\text{init}$ randomly sampled base poses in the scene and evaluate their corresponding scores. We then run $N^{(\text{BO})}_\text{iter}$ optimization rounds where each round proposes $N^{(\text{BO})}_\text{batch}$ new poses based on the acquisition function. The highest-scoring pose found among all iterations is selected as the final pose for executing the manipulation policy.
See implementation details of our method in the supplementary materials.

\section{The \name{} Framework}
\label{sec:framework}

To study policy mobilization in more depth, we propose the \name~framework, which includes metrics to quantify the difficulty of mobilizing a given policy (Section~\ref{sec:metric}), a suite of simulated mobile manipulation tasks to benchmark methods and policies for mobilization (Section~\ref{sec:suite}), as well as several visualization tools.

\subsection{Mobilization Feasibility Metrics}\label{sec:metric}
Given a policy for a task, we want to know how hard it is to mobilize it. We say a policy for a task is ``difficult to mobilize'' if running this policy on a mobile platform results in a large performance drop \textit{relative to} its in-distribution performance.
To quantify mobilization feasibility, we propose two distinct and complementary metrics as part of the \name{} framework.

\textbf{Spatial mobilization feasibility metric.} First, we quantify mobilization feasibility by measuring how robot base pose perturbations affect the task success of manipulation policies trained with fixed initial base poses. Specifically, we measure the task success $S_\pi(\sigma)$ of policy $\pi$ given Gaussian noise $N\sim(\mu=0, \sigma^2)$ applied to the base pose. We then fit an exponential curve using the following form: $
S_\pi(\sigma) = C_0 \cdot e^{-\gamma \sigma}$, where $C_0$ and $\gamma$ are parameters to be fitted.
From this fitted curve, we extract the \textit{spatial mobilization feasibility metric}: $\phi = \frac{\ln 2}{\gamma}$.
This metric $\phi$ can be interpreted as the deviation of the base pose in meters that causes the policy success rate to drop by half. The higher $\phi$, the easier it is to mobilize this policy as the policy performance decays more slowly.

\textbf{Visual mobilization feasibility metric.} The visual appearance of the tasks can also affect mobilization difficulty. For example, if a policy manipulates an object while barely seeing the object in view, it will be harder to find a good location for policy execution. To quantify this, we evaluate how relevant objects of the task are represented in the scene. Formally, we sample $N_v$ images of the scene at near-optimal poses and compute the average percentage $S_v$ of the images occupied by the objects of interest in the task.
The visual mobilization feasibility metric can be evaluated for methods that consider aligning the robot's viewpoint with the policy's capabilities during base placement. Naive baselines like LeLaN~\cite{hiroselelan} and VLFM~\cite{yokoyama2024vlfm} are not policy-aware, so this metric does not apply.

\subsection{Simulated Task Suite for Evaluating Policy Mobilization}\label{sec:suite}
\begin{figure*}[t!]
    \centering
    \includegraphics[width=\linewidth]{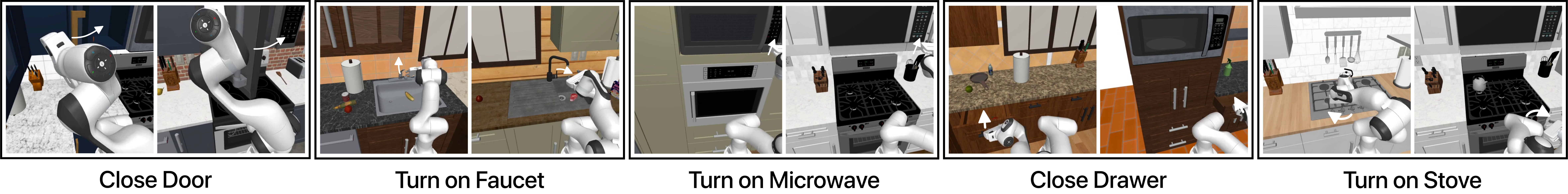}
    \vspace{-10pt}
    \caption{\textbf{A suite of simulated tasks for benchmarking performance of policy mobilization methods.} We pick five single-stage manipulation tasks from the RoboCasa benchmark~\cite{robocasa2024}: 
    (1) \textit{Close Door:} push or close the microwave or cabinet door using the robot gripper.
    (2) \textit{Turn on Faucet:} precisely move the faucet handle to turn on water.
    (3) \textit{Turn on Microwave}: push a small button on the microwave.
    (4) \textit{Close Drawer}: close an open drawer next to the robot.
    (5) \textit{Turn on Stove}: turn a specific knob out of six knobs on the stove.
    Here, we visualize two successful rollouts for each task. Note that episodes are executed in environments with different layouts and textures. 
    In the \textit{Close Door} task, the robot might encounter either a cabinet door or a microwave door. The \textit{Turn on Faucet}, \textit{Turn on Microwave}, and \textit{Turn on Stove} tasks require accurate robot placement to be executed correctly. In the \textit{Close Drawer} task, the drawer can be either on the left or right side of the robot.
    }
    \label{fig:sim_tasks}
    \vspace{-10pt}
\end{figure*}

Part of the \name{} framework is a suite of simulated mobile manipulation tasks based on RoboCasa~\cite{robocasa2024} that will allow us to study the policy mobilization problem and benchmark different methods. Our task suite includes five challenging tasks illustrated in Figure~\ref{fig:sim_tasks}.
For each task, we train a standard image-based imitation learning policy~\cite{shafiullah2022behavior} from 300 demonstrations. These demos are generated using MimicGen~\cite{mandlekar2023mimicgen} with procedurally generated textures in 5 training room layouts. In each task, we use a Franka Panda robot mounted on an Omron mobile base initialized from the same base pose. Each task comes with a language description and a horizon of 500 steps.
The goal of the benchmark is to deploy these policies into 5 held-out scenes with unseen room layouts and textures.
Note that since these tasks come originally from RoboCasa, the tasks and policies are not designed for the problem of policy mobilization. Therefore, this benchmark can serve as an unbiased challenge task suite for testing methods for policy mobilization.

\begin{figure*}[t!]
    \centering
    \includegraphics[width=\linewidth]{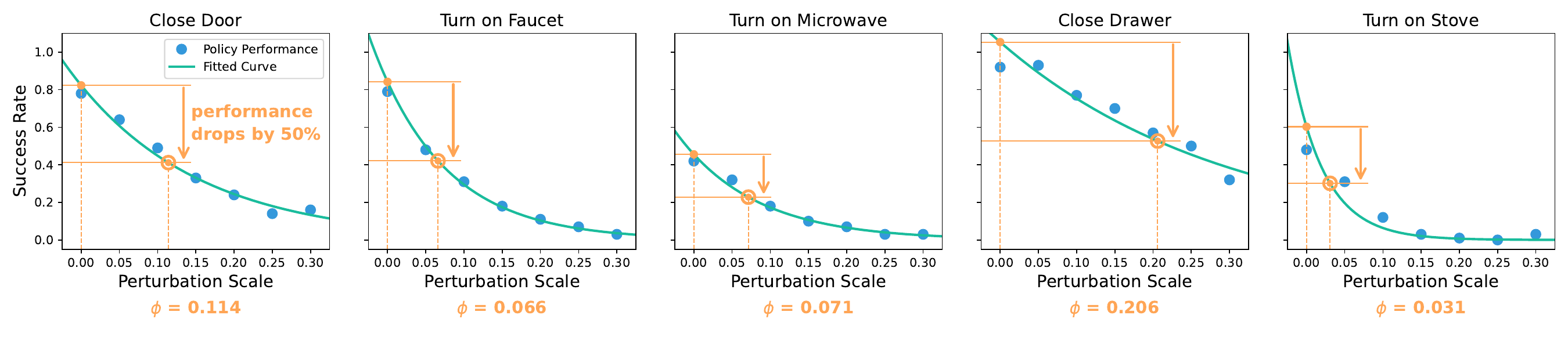}
    \vspace{-20pt}
    \caption{\textbf{Spatial mobilization feasibility.} We show the performance of running the given policy with Gaussian noise applied to the initial robot base pose at different standard deviations. The x-axis represents the standard deviation $\sigma$ of the applied noise (which we also call ``perturbation scale''). The y-axis represents the success rate of the policy evaluated for 150 episodes per blue dot. The green curves represent the fitted performance decay curve. Given the fitted curve, we can find the perturbation scale $\phi$ (in meters) that causes the policy success rate to drop by half. The smaller $\phi$, the harder it is to mobilize this policy since policy performance quickly decays with an increasing deviation from the in-distribution poses.}
    \label{fig:metrics}
    \vspace{-10pt}
\end{figure*}

\begin{figure*}[t!]
    \vspace{-6pt}
    \centering
    \includegraphics[width=0.7\linewidth]{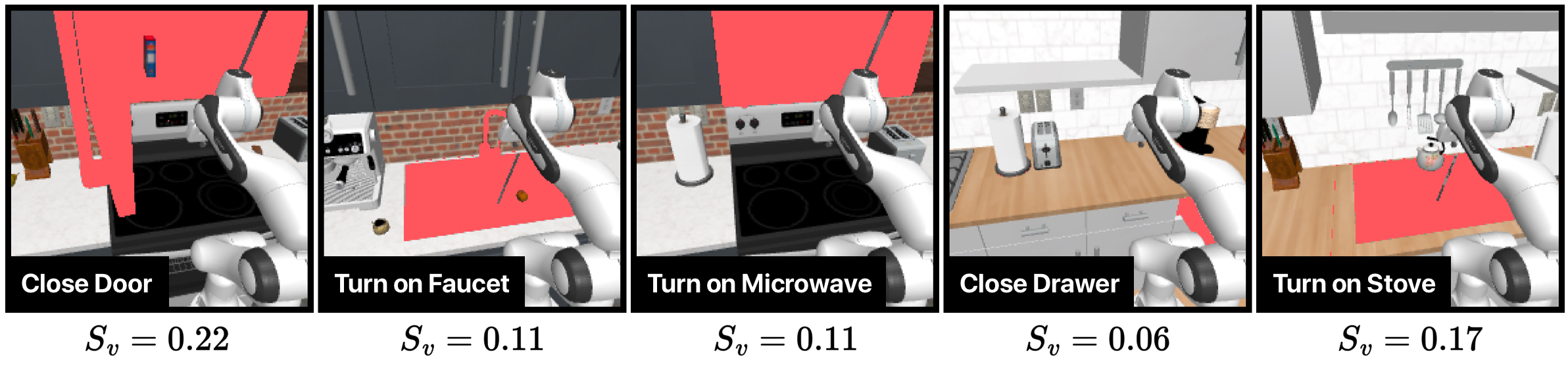}
    \vspace{-3pt}
    \caption{\textbf{Visual mobilization feasibility.} In each task, we evaluate how feasible it is to mobilize a robot learning policy from a visual perspective by computing on average, how much of the view is occupied by objects of interest. We denote this metric as $S_v$. This is important because the visual appearance of objects of interest will help policy mobilization methods decide a proper robot placement pose for policy execution. If the object of interest only takes up a tiny part of the scene, it is more difficult for a method to precisely position the robot such that the visual viewpoint is proper for policy execution.}
    \label{fig:vis_metric}
    \vspace{-16pt}
\end{figure*}

\textbf{Quantifying mobilization feasibility of benchmark tasks.}
We plot the spatial mobilization feasibility metrics of the policies in the five simulated kitchen tasks in Figure \ref{fig:metrics}. We observe that the most challenging task to mobilize based on our metric is the \textit{Turn on Stove} task ($\phi=0.031$). This reflects the difficulty to precisely turn a small knob which has limited contact area.
Figure~\ref{fig:vis_metric} shows the visual mobilization feasibility metric evaluated in the five environments. Results show that visual mobilization feasibilities do not necessarily match spatial mobilization feasibilities.
More specifically, the \textit{Close Drawer} task, which appears to be easy to mobilize spatially, turns out to be the most difficult in terms of the visual mobilization feasibility metric.
This is expected because even in spatially tolerant tasks, using proper visual features to find optimal policy initialization poses can be difficult. In experiments, we show how the mobilization metrics are correlated with the method performance.
\section{Experiments}
\label{sec:exp}

We aim to investigate how our policy-aware method performs compared to other policy-aware and non-policy-aware baselines, if our mobilization feasibility metrics are predictive of method performance, and if our method can be reliably deployed in real-world settings.

\textbf{Baselines.} We divide baselines into two categories. The first type of baselines navigates to the object of interest \textit{without considering the manipulation policy's capabilities}. These methods are \textit{not policy-aware}. Representative methods include: (1) \textit{LeLaN}~\cite{hiroselelan}, a language-conditioned navigation model trained from large-scale videos. Since the pre-trained LeLaN model is trained on only real-world data, we fine-tune the pre-trained LeLaN model with 7.5k episodes of in-domain, randomized navigation data in simulation. At test time, we run LeLaN for a fixed number of steps to approach the object while using the scene point cloud for collision checking, and then execute the target policy. (2) \textit{VLFM}~\cite{yokoyama2024vlfm}, a zero-shot navigation model that explores the scene, uses a ground-truth detector to locate the object, selects a point on its unprojected point cloud, and navigates to it using RRT and the scene point cloud for collision checking. As shown in our experiments, these types of methods often fail to find suitable base poses as they are not taking manipulation policy performance into account, resulting in failed policy executions.

The second type of baselines is \textit{policy-aware} and leverages large-scale data to connect navigation with manipulation. 
A representative method is
\textit{BC w/ Nav}: a Behavior Transformer~\cite{shafiullah2022behavior} trained to jointly perform navigation and manipulation using combined demonstrations. We provide 1,500 demonstrations per task, $5\times$ the number of demos used compared to our base manipulation policy. We show in experiments that these methods generalize poorly to unseen layouts despite extensive data collection. See supplementary materials for baseline details.

\subsection{Simulation Experiments}
\label{sec:sim_exp}

\textbf{Metrics.}
In addition to reporting the success rate of the baselines and our method, we also report \textit{Manipulation Policy Performance}, which is the success rate of the original, non-mobile robot policy per task without mobilization. Note that the success rates of the base policies vary significantly across tasks. We report the success rate and standard deviation across all tasks.

\textbf{Evaluation setup.} 
In each task, we evaluate our method and competing baselines in 10 scenes with distinct layouts and textures. For each of the 3 random seeds, we run 50 evaluation episodes, 5 episodes in each scene. 
In every episode, the robot is randomly placed in the scene with the target object in view. The method needs to navigate and then successfully execute the manipulation policy.
We report the mean and standard deviation over the policy execution success rates. 
Each result value includes a total of 150 evaluation episodes.

\begin{figure*}
    \centering
    \includegraphics[trim=0 0 0 0,clip,width=0.9\linewidth]{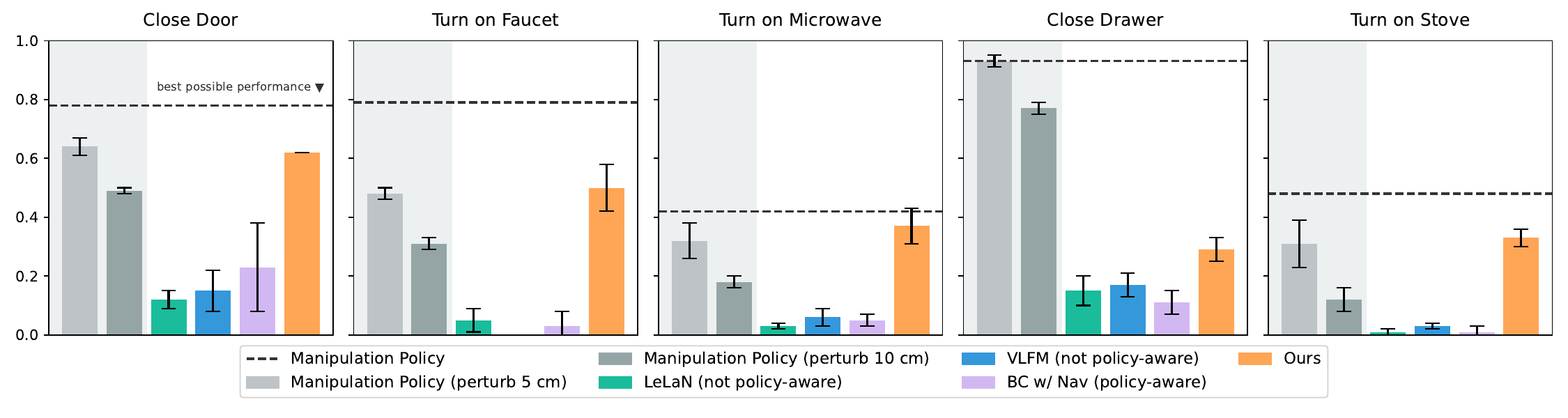}\vspace{-6pt} 
    \caption{\textbf{Simuation experiment results.} We evaluate {\color{my_green}\textbf{LeLaN (non-policy-aware)}}, {\color{my_blue}\textbf{VLFM (non-policy-aware)}}, and {\color{my_purple}\textbf{BC w/ Nav (policy-aware, 5x training data)}} as well as {\color{my_orange}\textbf{our proposed method}} in the 5 simulated tasks. For each task, we evaluate the success rate of task execution for 3 seeds and 50 episodes per seed. The plotted bars represent the mean and standard deviation over the results of the 3 seeds. We also report the {\color{my_gray}\textbf{manipulation policy performances}} with and without perturbation of the starting base pose. In four of five tasks, our method matches up with the performance of the manipulation policy when executed with a 5 cm perturbation at initialization.}
    \vspace{-2pt}
    \label{fig:sim_main}
\end{figure*}

\begin{wrapfigure}{r}{0.4\textwidth}
    \vspace{-13pt}
    \centering
    \includegraphics[width=\linewidth]{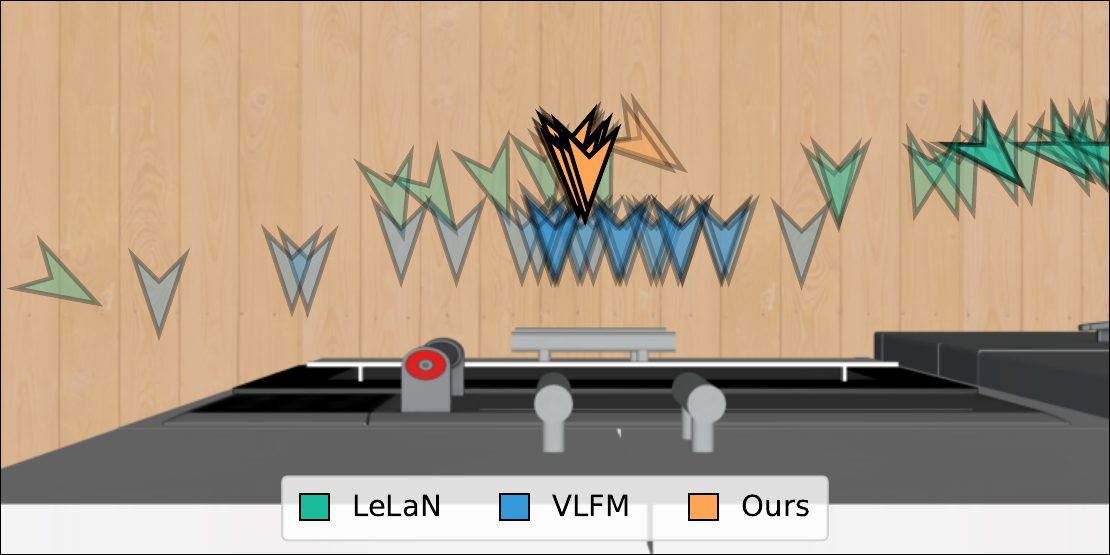}
    \vspace{-13pt}
    \caption{\textbf{Visualization of qualitative performances in simulated experiments.} This plot shows achieved navigation end poses of the robots produced by multiple methods in a sample scene in the \textit{Turn on Microwave} task. The background plots the top-down map of the scene where the robot needs to run the designated policy. Each arrow shows a robot pose achieved by a method after navigation. The colors of the arrows represent method names. Semi-transparent arrows indicate failures and opaque arrows indicate successes. In the plot, our method consistently reaches the microwave door with the appropriate heading, while baselines fail to do so.}
    \label{fig:sim_qual}
    \vspace{-20pt}
\end{wrapfigure}

\textbf{Results.}
We present quantitative results for our method in Figure~\ref{fig:sim_main}. 
\textit{LeLaN (non-policy-aware)} and \textit{VLFM (non-policy-aware)} perform poorly because they do not consider which robot poses enable a successful policy rollout (see Figure~\ref{fig:sim_qual} for qualitative results).
\textit{BC w/ Nav (policy-aware)} performs poorly in all tasks, showing that it is difficult to train a working mobile manipulation policy that generalizes to out-of-distribution room layouts.
Our method outperforms all baselines in all tasks, achieving better performance in navigating to a pose suitable for executing the downstream policy, resulting in higher task success.

\textbf{Relation to mobilization feasibility metrics.} Baselines that are not policy-aware perform worst in \textit{Turn on Faucet}, \textit{Turn on Stove}, and \textit{Turn on Microwave}. These are also tasks that have the lowest spatial mobilization feasibility scores (Section~\ref{sec:metric}). This suggests that poor spatial tolerance of a manipulation policy makes it challenging for a non-policy-aware method to perform well. Our method's performance also correlates with the visual mobilization feasibility metric. It matches or exceeds \textit{Manipulation Policy (Perturb 5 cm)} across tasks, except \textit{Close Drawer}, which has low visual feasibility. As our method relies on visual cues provided by the target objects, this explains the performance drop.

\begin{wraptable}{r}{0.5\textwidth}
\vspace{-2pt}
\centering
\scriptsize
\setlength{\tabcolsep}{4pt}
\begin{tabular}{lcc}
\toprule
 & \textbf{Close Door} & \textbf{Turn on Microwave} \\
\midrule
Ours & \textbf{0.62 $\pm$ 0.00} & \textbf{0.37 $\pm$ 0.06} \\
Ours w/o feature descriptor & 0.41 $\pm$ 0.05 & 0.07 $\pm$ 0.08 \\
Ours w/ policy encoder & 0.11 $\pm$ 0.07 & 0.00 $\pm$ 0.00 \\
\bottomrule
\end{tabular}
\caption{\textbf{Simulation ablation experiment results.}}
\label{tab:sim_ablation}
\vspace{-10pt}
\end{wraptable}

\textbf{Ablations and analysis.}
We conduct ablation studies in two simulation environments, \textit{Close Door} and \textit{Turn on Microwave}.
We consider the following ablated methods: 
(1) \textit{Ours w/o Feature Descriptor}: this ablated method uses a DINO embedding vector instead of the DINO dense feature descriptor in $K_\text{id}$ to represent information in an image.
(2) \textit{Ours w/ Policy Encoder}: this method uses the ResNet-18~\cite{he2016deep} encoder of the given policy to represent an image in $K_\text{id}$.

We present the results of this ablation in Table~\ref{tab:sim_ablation}. \textit{Ours w/o Feature Descriptor} underperforms compared to our method, as the 1D DINO embedding lacks spatial cues necessary for distinguishing camera views. \textit{Ours w/ Policy Encoder} also performs poorly, possibly because it focuses on task-relevant scene features, ignoring view-dependent cues. Since the policy is trained only with action-related losses, it may not capture the information needed to guide robot placement.

We also design visualization tools to show the qualitative performance of competing methods. In Figure~\ref{fig:sim_qual}, we show top-down maps of the simulated benchmark environment overlaid with arrows to illustrate the navigation end poses achieved by different methods. See more ablations and analysis in the supplementary materials.

\vspace{-5pt}
\subsection{Real Robot Experiments}

\begin{wrapfigure}{r}{0.4\linewidth}
    \centering
    \vspace{-12pt}
    \includegraphics[width=\linewidth,clip]{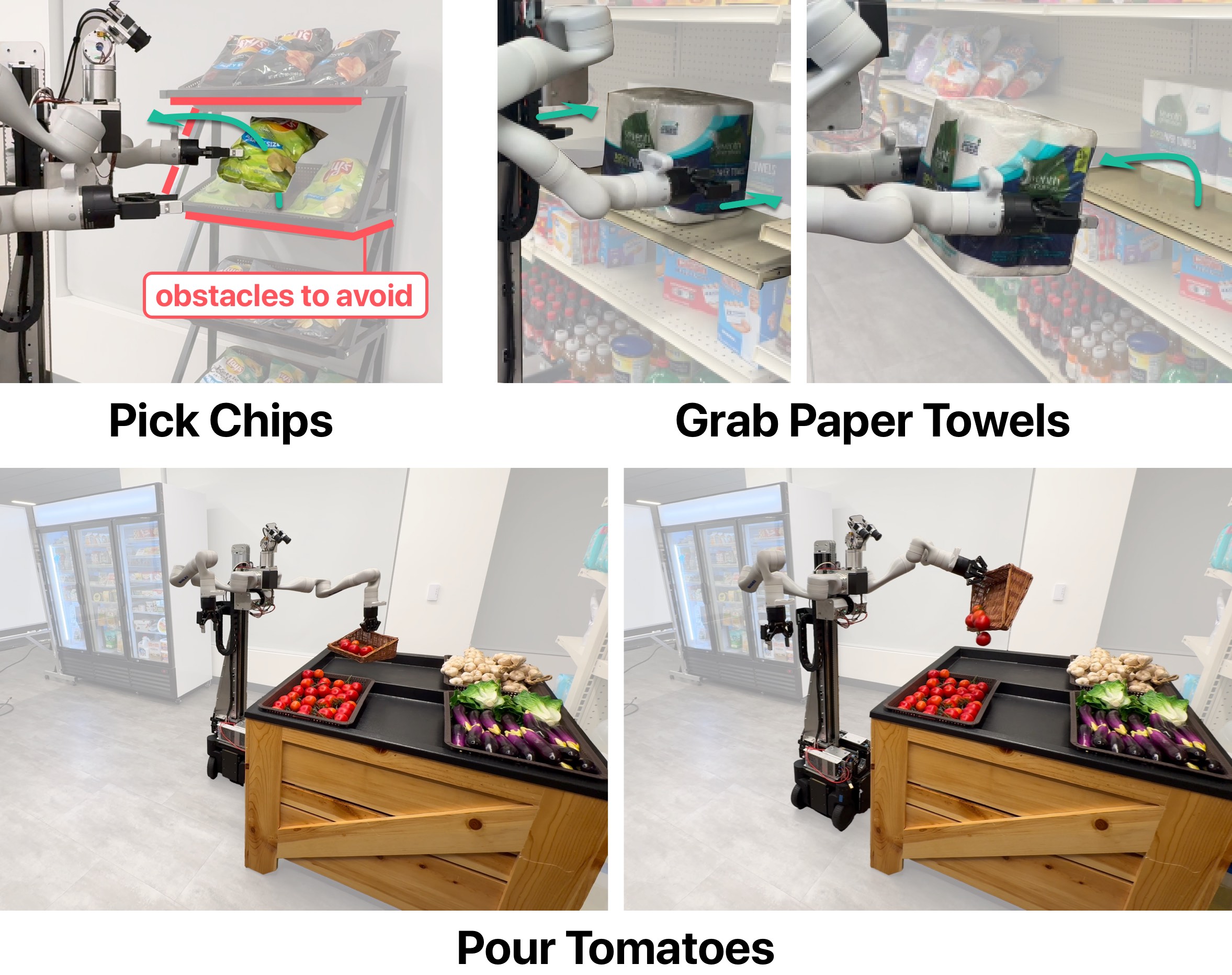}
    \vspace{-15pt}
    \caption{\textbf{Real-world tasks.} We pick three tasks that involve diverse objects and contact modes. (1) \textit{Pick Chips}: the robot retrieves a deformable bag of chips from a shelf.
    (2) \textit{Bimanual Grab Paper Towels}: the robot squeezes a pack of paper towels on a shelf between the two arms using non-prehensile manipulation, lifts it up, and pulls it off the shelf.
    (3) \textit{Pour Tomatoes}: the robot picks up a tray of tomatoes on the left side of a produce bin, moves rightward via base motions, and pours the tomatoes from the tray into a basket. See videos on our \href{https://mobipi.github.io/}{website}.}
    \label{fig:real}
    \vspace{-15pt}
\end{wrapfigure}

To validate that our method can be effectively deployed in real-world settings, we show a series of real robot experiments where a mobile robot performs various manipulation tasks in a mock grocery store (see Figure~\ref{fig:real}).

\textbf{Data collection and training.} We collect 30 human-teleoperated demonstrations for the \textit{Pick Chips} and \textit{Grab Paper Towels} task and 50 for the \textit{Pour Tomatoes} task.  We use a Basler camera mounted on the mobile robot to collect visual information and record both the images and the base and arm poses at a frequency of 10 Hz. We use the collected data to train a Diffusion Policy~\cite{chi2023diffusionpolicy}. See supplementary materials for more robot setup and training details.

\textbf{Baselines and rollouts.}
We initialize the robot at 10 random poses located 0.5 to 1.5 meters from the target policy execution pose, with the constraint that the target scene must remain within the robot's field of view. 
From each initialization, we execute both our proposed method and the baseline methods. For real-world experiments, we compare against one policy-aware and one non-policy-aware baseline. As the policy-aware baseline, we use \textit{BC w/ Nav}. To construct a strong non-policy-aware baseline, we introduce a \textit{Human} baseline: participants are asked to manually drive the robot to the location they believe will maximize task success, without being given any information about how the manipulation policy was trained. See details about this baseline in the supplementary materials.

\begin{figure*}
    \centering
    \includegraphics[trim=0 0 0 0,clip,width=\linewidth]{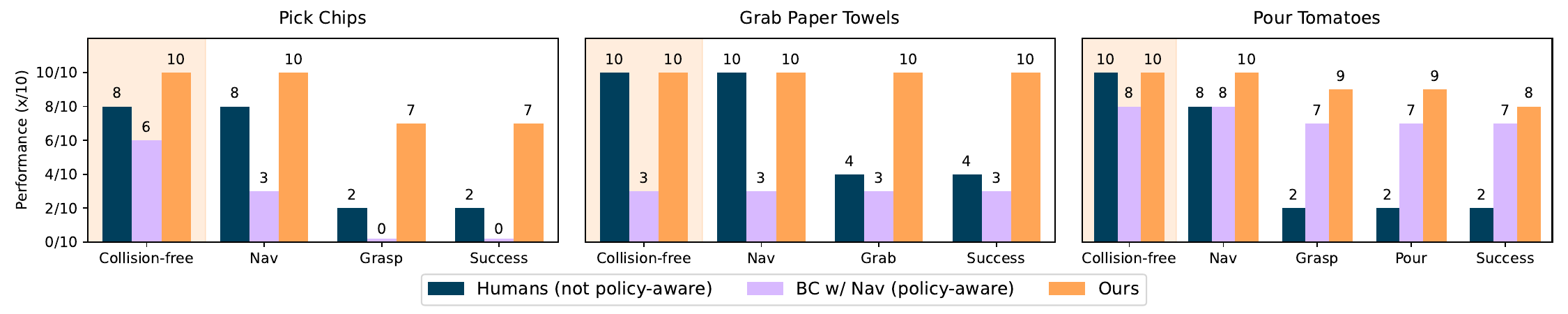}\vspace{-6pt} 
    \caption{\textbf{Real-robot experiment results.} For each task and method evaluated, we track the incidence rates of certain performance milestones (i.e., navigation success, grasp success, full success). We also track the percentage of collision-free episodes. We consider a navigation attempt successful if the robot drives to a pose that not only faces the target object, but also is within a distance of 50~cm from the ideal target pose. A grasp success is defined as when the robot stably grasps the object of interest (i.e. the chip bag, the paper towels, or the tray of tomatoes), and full success is counted only if the entire task is completed.}
    \label{fig:real_robot_results}
    \vspace{-15pt}
\end{figure*}

\textbf{Quantitative results.}
Quantitative results from the real-robot experiments are shown in Figure~\ref{fig:real_robot_results}. The \textit{Human} baseline displays an interesting pattern: although participants successfully navigate the robot near objects of interest, their lack of knowledge about the manipulation policy's capabilities often results in suboptimal base poses, leading to low success rates. The \textit{BC w/ Nav} baseline achieves higher performance but frequently positions the robot inappropriately--too close or poorly oriented realtive to the target object--resulting in failed manipulation attempts. In contrast, our method outperforms both baselines, reliably identifying base poses that enable successful policy execution.

\textbf{Qualitative results.}
\begin{figure*}
    \centering
    \includegraphics[trim=0 0 0 0,clip,width=\linewidth]{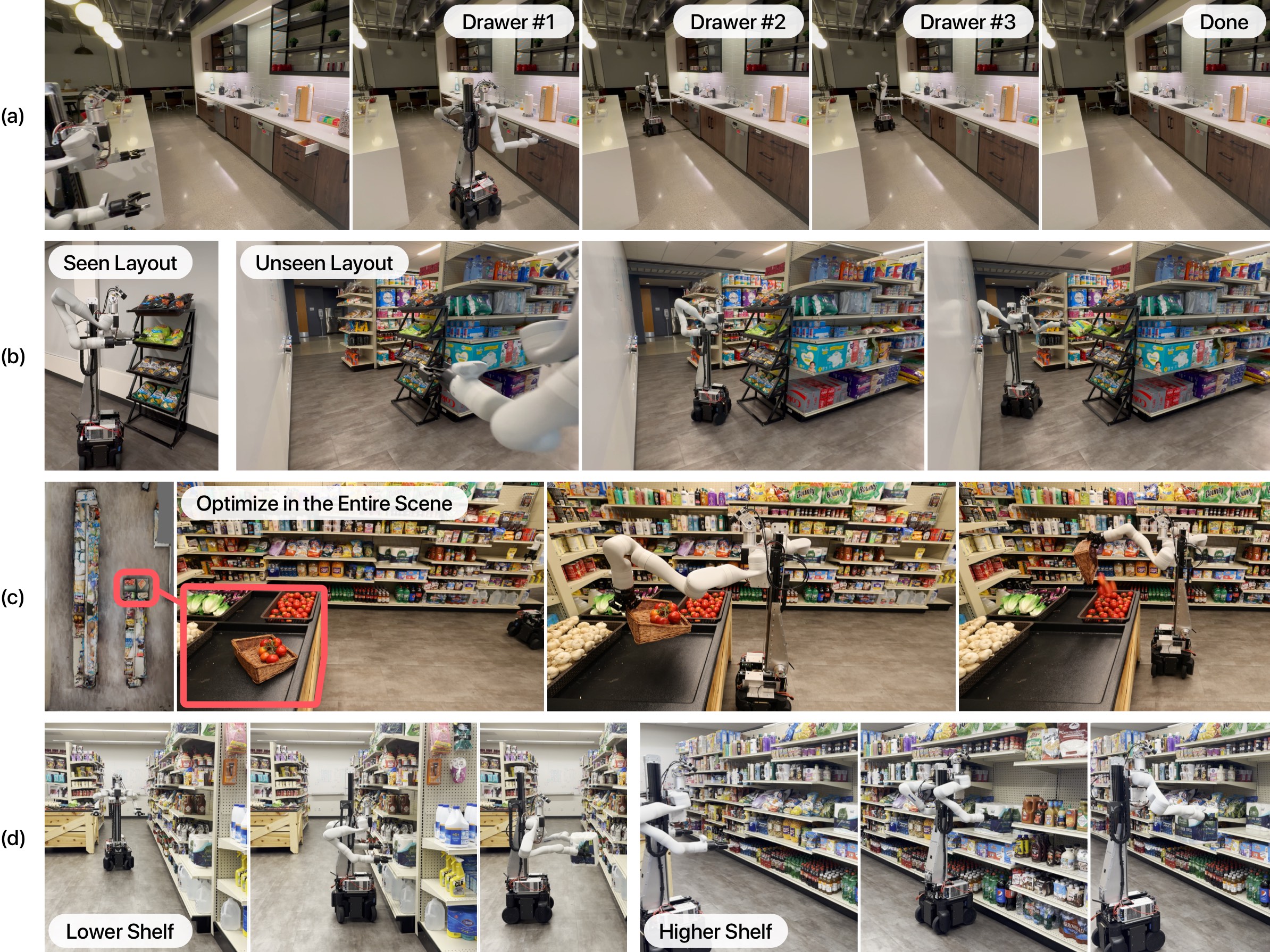}
    \caption{\textbf{Real-world qualitative results.} (a) In a real-world kitchen scene, we show that our method can chain a sequence of drawer-closing policy executions. (b) In the \textit{Pick Chips} task, we show that our method achieves zero-shot generalization to unseen scene layouts. (c) In the \textit{Pour Tomatoes} task, we show that our method is capable of operating in large spaces that the policies have not explored during training. The top-down image on the left shows the space in which our method performs its optimizations. This image is generated via a scan of the scene produced by the PolyCam iPhone app and is for visualization purposes only. (d) In the \textit{Bimanual Grab Paper Towels} task, we show that our method can be easily extended to optimize for robot and camera height as well.}
    \vspace{-10pt}
    \label{fig:real_qualitative}
\end{figure*}
Aside from quantitative results, we also demonstrate real-world deployment of our method in a variety of interesting ways (see Figure~\ref{fig:real_qualitative}).

First, in a real kitchen setting, we show that by querying our method multiple times, our method is capable of finding proper initial robot poses for multiple drawer closing skill executions in the scene, thus unlocking manipulation skill sequencing in the wild. During each round of base pose optimization, we avoid sampling poses that are close to the best poses found by previous optimization rounds to avoid running the drawer closing skill at the same location more than once. We assume that we know how many times the drawer closing skill needs to be executed. 

Second, we show that our method generalizes zero-shot to unseen scene layouts in the \textit{Pick Chips} task. This is accomplished by simply running our method in the new scene, without any change in the policy or hyperparameters of our method.

Third, we demonstrate in the \textit{Pour Tomatoes} task that our method is capable of operating in a large scene. We create a 3D Gaussian Splatting model of the entire grocery store scene, which spans $6\times 10$ meters in size. We then query our method for the optimal starting robot base pose. Our method successfully completes navigation and manipulation policy execution in this large scene. 

Finally, we show that our method can be easily extended to adapt to unseen object heights. To achieve this, we simply add one more delta robot height parameter to the optimization process and command the robot to move to the target height instructed by our method before starting policy execution. We find that our method can flexibly determine the proper height to run the \textit{Bimanual Grab Paper Towel} policy when the object is placed on shelves with different heights.
\section{Conclusion}

We introduced \name{}, a framework for studying policy mobilization. We formalized the policy mobilization problem, proposed a new method for solving this problem, and introduced an evaluation suite with simulation benchmarks, visualizations, and baselines. We demonstrated our method in simulated kitchen environments as well as a real-robot setup in a mock-grocery environment. This work provides a robust pathway for utilizing countless non-mobile robot policies on mobile platforms, enabling robots to perform intricate tasks in larger and more diverse environments.

\section{Limitations and Discussions}

While our method reliably identifies suitable initial robot base poses for executing non-mobile manipulation policies on a mobile platform, it does have several limitations.

\textbf{Dynamic scene changes.} 
Our approach assumes a static environment during policy execution and does not support dynamic scene updates. This limitation becomes apparent when chaining multiple policies in the same workspace: after executing one policy, changes in the environment are not reflected in the scene model, potentially leading to degraded performance in subsequent steps. Addressing this limitation would require integrating a dynamic neural rendering model that can be updated in real-time as the robot interacts with the scene~\cite{wu20244d, zheng2025gstargaussiansurfacetracking}.

\textbf{Access to training data.} 
Another potential limitation is the requirement to access the manipulation policy's training data. However, we emphasize that policy mobilization—by design—requires some degree of policy awareness, which inherently necessitates access to policy-related information. For example, the \textit{BC w/ Nav} baseline achieves this by training a navigation model on additional task-specific data, which can be expensive to collect. In contrast, our method relies only on the initial observation frames from the demonstration dataset, making it less data-intensive. Nevertheless, future work could explore reducing this requirement further—e.g., by sampling a smaller subset of initial frames or, in the absence of training data, using a few successful rollouts of the policy and extracting initial observations from those episodes.

\textbf{Supporting multi-task policies.}
Our current study focuses on single-task manipulation policies. Extending policy mobilization to multi-task settings, where the policy is conditioned on goal images or language instructions, is a promising direction for future work.

\textbf{Fixed base-to-camera transform.}
Although the policy mobilization problem does not impose strict assumptions on the base-to-camera configuration, our current implementation assumes a fixed transformation between the robot base and its onboard camera. A natural extension of our approach would be to jointly optimize over both the base pose and camera configuration. This could be achieved by making the robot visible from one of the camera views and running the same iterative optimization algorithm with the camera viewpoint as an additional optimization parameter within the existing framework. Exploring this direction is an exciting direction for future research.

\newpage
\acknowledgments{This work was completed when Jingyun was an intern at Toyota Research Institute. This work was supported by the Toyota Research Institute.

We thank members of the Mobile Manipulation Team at Toyota Research Institute for building the robot used in this work and maintaining related infrastructures. We thank members of the Stanford IPRL lab, Leonidas J. Guibas, Sergey Levine, Shuran Song, Zi-ang Cao, Yufei Ding, Ray Song, Chen Wang, and Haoyu Xiong for their helpful discussions. We thank Cherie Ho, Marion Lepert, and Kevin Lin for their help during the rebuttal.}

\printbibliography

\newpage
\setcounter{section}{0}
\renewcommand\thesection{\Alph{section}}
\renewcommand\thesubsection{\thesection.\arabic{subsection}}

\title{Supplementary Materials}
\makesupptitle

\section{Frequently Asked Questions}

\subsection{Motivation and Alternative Approaches}

\textbf{Why should I care about policy mobilization? Why not simply train a mobile manipulation policy from a large dataset?} Indeed, one approach to learning mobile manipulation is to train a policy from a dataset that includes both navigation and manipulation data. However, training such a policy typically requires large amounts of training data, since the policy needs to not only generalize to different navigation and manipulation scenarios but also seamlessly coordinate navigation and manipulation. We showed in our simulation experiments that the \textit{BC w/ Nav} baseline fails to perform well in unseen room layouts despite learning from $5\times$ more training data (see Section~\ref{sec:sim_exp}). Compared to training an end-to-end mobile manipulation policy, our policy mobilization framework provides a more data-efficient approach to learning mobile manipulation.

\textbf{Why not train a mobile manipulation policy in simulation and transfer to the real world?} Training a mobile manipulation policy in simulation from 3DGS requires creating an interactive and physically accurate simulation from the 3DGS model, which is an unsolved problem.

\textbf{Why not consider an image-goal navigation method or further achieve fine-grained navigation~\cite{meng2025aim}?} Image-goal navigation won't easily apply because it is unclear what goal image to use. For example, in sim, the policy is deployed in novel environments with out-of-distribution layouts and textures, and it is trained on data from multiple diverse scenes. This means that the test-time scene does not match any scene captured during policy training. A similar argument applies to fine-grained navigation: one needs to first search for \textit{where to navigate to} before running navigation.

\subsection{Method}

\textbf{How sensitive are DINO dense descriptors to viewpoint changes?} In the early stage of the project, we compared several feature representations, including DINO dense descriptors, DINO features, policy encoder features, ResNet-50~\cite{he2016deep}, and DeepLab v3~\cite{chen2017rethinking}. We empirically found that the DINO dense descriptor performs the best in finding an in-distribution image out of a set of similar-looking images rendered from various viewpoints. In the paper, we showed quantitative ablation results where our method outperforms variations of our method where the DINO dense descriptors are replaced with flat DINO features and policy encoder features. These observations coincide with findings in prior work~\cite{amir2021deep, hadjivelichkov2023one} that DINO dense descriptors carry rich, well-localized semantic and positional information. Also, note that our method is compatible with any improved visual feature representation that emerges in the future.

\begin{wrapfigure}{r}{0.3\textwidth}
    \vspace{-27pt}
    \centering
    \includegraphics[width=\linewidth]{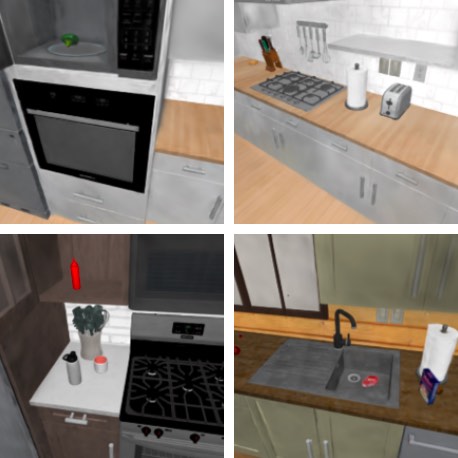}
    \vspace{-12pt}
    \caption{\textbf{Sample 3D Gaussian Splatting renders.} As seen in the samples, our 3DGS models synthesize novel views that are imperfect. Nevertheless, our method maintains good performance.}
    \label{fig:sample_renders}
    \vspace{-24pt}
\end{wrapfigure}

\textbf{How does the approach handle distractors?} Our sim setup includes unseen test objects. We find that our method, which utilizes DINO dense descriptors to score robot poses, is capable of ignoring these irrelevant objects.

\textbf{Can the method handle imperfect scene reconstruction?} Yes. Our learned 3DGS models have artifacts and surface color inconsistencies (see Figure~\ref{fig:sample_renders}), yet the method still performs well.

\textbf{The proposed method builds a 3D Gaussian Splatting model. Does it scale to larger, more complex scenes?} In simulation, we showed that our method effectively finds optimal base positions in kitchen scenes as large as $36 m^2$. This proves that our method is useful in room-scale settings. To make our method useful in a multi-room setting, one can consider designing a system that first navigates to the correct room and then executes our method. In the real world, another bottleneck is building a high-quality 3D Gaussian Splatting model. Prior work~\cite{yu2024language} studied building 3DGS models for complex scenes by incrementally registering images captured using a multicamera system to minimize drift during scene mapping. Integrating methods like this into our framework is a clear direction for future work.

\textbf{Is the visual feasibility metric general to the problem of policy mobilization?} Our visual metric applies to methods that consider aligning the robot's viewpoint with the policy's capabilities during base placement. Naive baselines (e.g. LeLaN, VLFM) for policy mobilization are not policy-aware, so this metric does not apply.

\textbf{Why does the proposed method use L2 distance in latent space as the in-distribution score rather than a distribution score (e.g. compute the exact likelihoods such as normalizing flows (NFs) as in~\cite{feng2023topology})?} 
While we could indeed search among more frames in the training dataset, doing so is computationally more expensive. We empirically found that only using the initial frames led to good performance.
Using the L2 distance on DINO features is a simple, off-the-shelf, and empirically effective \textit{proxy} for distribution matching that does not require training complex density models like normalizing flows (NFs), which require additional training, careful tuning, and might overfit on small datasets. We also note that the task suite we developed in sim can be used to develop alternative score functions that utilize more than just initial frames or employ flow-based methods.

\subsection{Experiments}

\textbf{The proposed method requires access to training data. Are there ways to relax this assumption?} We explained in the paper that, to appropriately link navigation and the execution of a manipulation policy, a policy mobilization method needs to be \textit{policy-aware}. This means that the policy mobilization method needs to have some information about the manipulation policy. Our proposed method assumes access to the initial observations of the training demonstrations. This is usually easy to obtain in open-source datasets and models. We mentioned in the Limitations section that there are ways to relax this assumption if the dataset is too large or not publicly available. For example,  you could sample a subset of demonstration episodes or rolling out the policy to collect some successful episodes. Note that the baselines also assume access to in-distribution data to different extents: \textit{BC w/ Nav} requires collecting additional in-distribution data for policy learning, and \textit{LeLaN} is fine-tuned with in-distribution navigation data.

\textbf{The proposed method has access to the 3D model of the test-time scene. How did you ensure a fair comparison to baselines that natively do not assume this information?}
To ensure a fair comparison, we provide the \textit{LeLaN} and the \textit{VLFM} baselines with ground-truth 3D models of the scene for collision detection. This detail is also mentioned in Section~\ref{sec:exp} of the main paper. The \textit{BC w/ Nav} cannot take into account a 3D model of the test-time scene. Therefore, we do not provide it.

\section{Additional Ablation Results}

\begin{wraptable}{r}{0.5\textwidth}
\vspace{-10pt}
\centering
\scriptsize
\setlength{\tabcolsep}{4pt}
\centering
\begin{tabular}{lcccc}
\toprule
  & \textbf{Ours}  & \textbf{w/o $K_\text{id}$} & \textbf{w/o $K_\text{col}$} & \textbf{w/o $K_\text{obj}$} \\ \midrule
Mean Success Rate & 0.422 & 0.006                  & 0.408                   & 0.344                    \\ \bottomrule
\end{tabular}
\caption{\textbf{Ablations on hybrid score function.} We compute the average success rate of each method across all 5 tasks, with a total of 750 evaluation episodes for each number shown in the paper.}
\label{tab:abl_hybrid_score}
\vspace{-13pt}
\end{wraptable}

To study the individual importance of each element in the hybrid score function, we add an ablation study where we compare our method with variants of our method without one of the three hybrid score components. In Table \ref{tab:abl_hybrid_score}, we show that each component of the hybrid score function contributes to the final performance of our proposed method.

\section{Method Details}

\subsection{Hyperparameters}

When training Gaussian Splatting models, we use an image size of $512 \times 512$ in all simulated experiments and $1280 \times 1024$ in all real robot experiments. In simulation, since we know the ground truth camera poses without noise, we turn off the camera pose optimizer in \texttt{Nerfstudio}. In real robot experiments, since robot odometry drifts over time, we turn on the camera pose optimizer. We observe that turning the camera pose optimizer on improves the quality of the trained Gaussian Splatting model in the real world.

During hybrid score computation, we use an image size of $224 \times 224$ to match the input dimensionalities of the DINO encoder. When computing $K_\text{id}$, we use $k = 5$ for the KNN model. When computing $K_\text{obj}$, we query the MiniCPM-v2 model with the image and a prompt that says ``Is [object name] in the image? Answer exactly `yes' or `no'.'' We then return positively if the answer includes the substring ``yes''.

In all simulated experiments, we use $N^{(\text{BO})}_\text{init} = 2500$, $N^{(\text{BO})}_\text{iter} = 100$, and $N^{(\text{BO})}_\text{batch} = 5$. In real robot experiments, we use $N^{(\text{BO})}_\text{init} = 1000$. We use the UCB acquisition function with $\kappa = 1.96$ for simulated experiments and $\kappa = 0.5$ for real robot experiments.

\subsection{Implementation Details}

Our method takes as input one image at a time, but can be trivially extended to a set of input images. This is useful when the robot is equipped with multiple cameras for example at the wrist or from an additional third-person viewpoint. To compute a predicted score $K(p)$ in this case, we evaluate the  $K_\text{id}$ and $K_\text{obj}$ scores for each view, take the max, and then compose with the $K_\text{col}$ score to obtain the final score.

In simulation experiments, for all tasks but the \textit{Close Drawer} task, we only use one camera view to compute $K(p)$ since the right camera view is always the best camera view. However, in the \textit{Close Drawer} task, we found the need to use both camera views since the drawer might be either on the left or right side of the robot, resulting in different views being optimal for deciding the $K(p)$ score for different scenes and setups.

We also add minor modifications to our method to deal with language-conditioned policies. When computing the score function $K(p)$ in one episode, one can simply filter the training images $\{o_1^k\}_{k=1}^{N_{DS}}$ to include only images that have matching language descriptions of the current task description. In this way, the resulting score will be computed corresponding to the task the agent is currently performing.

We run our method on a Linux workstation equipped with an NVIDIA RTX 4090 GPU. Training a 3D Gaussian Splatting model on this machine takes approximately 15 minutes. Running a round of sampling-based robot pose optimization with Bayesian Optimization takes approximately 6 minutes.

\section{Baseline Details}

\textbf{LeLaN.} In our experiments, we assume that the robot initializes at a base position where it can navigate in a straight line to the target object without collision. Therefore, we used a pre-trained LeLaN~\cite{hiroselelan} checkpoint without the collision loss objective. We tried the pre-trained checkpoint in simulation and found it to perform poorly. To improve the robustness of the baseline, we fine-tune the pre-trained checkpoint with a randomized navigation dataset. For this dataset, we collected 300 demonstrations in each training layout and task, totaling $300 \times 5 \times 5 = 7,500$ navigation demonstrations. Each demo has a different, procedurally generated scene texture. In each demonstration, we retrieve the ground-truth target object position, add an offset to this position so we find the nearest robot pose in front of the object without collision, and add random noise to the position and rotation of this pose. As a result, we obtain a dataset of non-policy-aware navigation data where the robot initializes in training scenes and navigates to objects that the robot will encounter at test time. We fine-tune LeLaN on this generated dataset for 20 episodes before deploying it into each test-time environment. We only fine-tune a single LeLaN checkpoint to be used for all tasks.

\textbf{VLFM.} We integrated VLFM~\cite{yokoyama2024vlfm} into our framework with several key enhancements. Images and segmentation masks from RoboCasa~\cite{robocasa2024} were sent to VLFM, which generated high-level navigation commands such as \texttt{TURN LEFT}, \texttt{TURN RIGHT}, and \texttt{STOP}, based on its algorithm using pretrained feature extractors and object detectors to build semantic value maps for its exploration phase. Following exploration, we replaced VLFM’s pretrained navigation module (originally trained in Habitat~\cite{savva2019habitat}) with our own implementation of RRT, utilizing a point cloud representation of the scene for collision checking. Rather than navigating to the closest point on the target object, we improved accuracy by selecting the median point within the object’s point cloud. Additionally, we enhanced the performance of the pretrained MobileSAM~\cite{zhang2023faster} detector by incorporating RoboCasa’s ground truth masks, which significantly improved segmentation accuracy.

\textbf{BC w/ Nav.} In simulation, we collect 1,500 demonstrations per task, where each demonstration includes navigation and manipulation chained together. This is 3$\times$ more demos than the number of demos used for training the pure manipulation policy. Note that each demo also has a longer horizon than the demos used to train a manipulation policy because of the additional navigation steps. In real-robot experiments, we collect the same number of episodes to train \textit{BC w/ Nav} as our method. We use the same algorithm, hyperparameter setting, and number of epochs to train this baseline as the manipulation policy used for mobilization.

\section{Real Robot Experiment Details}

\textbf{Robot setup.}
In all real robot experiments, we use a custom mobile robot with a wheeled mobile base, a prismatic lift that can be used to adjust the height of the robot arms, and two Kinova Gen3 7DoF arms mounted on the left and right side of the robot. The robot also has two Basler cameras mounted inbetween the two Kinova arms. 
In total, the robot has 22 degrees of freedom (3 for the base, 1 for the lift, 7 for each arm, 1 for each gripper, and 2 for the head joints used for adjusting viewing angles of the Basler cameras). During policy execution, we only control the base and arms of the robot and keep head and lift joints fixed. In the \textit{Pick Chips} and \textit{Pour Tomatoes} tasks, only the left arm will have non-zero actions, though we keep the action space the same by zero-padding right arm actions.

\textbf{Policy learning.} The base policies in our real robot experiments take as input an image of the camera view and the robot end-effector poses, and outputs desired robot arm position and rotation velocities. To make a lightweight policy, we downsample the RGB images to $256 \times 320$ before passing them into the policy. Similar to the original Diffusion Policy paper, we use an observation horizon of 2 steps, a prediction horizon of 16 steps. Since there are latencies between observation retrieval and execution of the predicted actions, we perform action skipping so that the robot executes the actions that the policy expects it to run. After action skipping, we run all remaining predicted actions before performing another round of action inference. We use the DDPM noise scheduler with 100 diffusion steps during training. We train all policies for 1,000 epochs before deploying them onto the robot.

\textbf{Rollout.} We develop a multiprocessed pipeline for evaluating trained Diffusion Policies~\cite{chi2023diffusionpolicy} in real robot setups. An inference process repeatedly receives observations from the Basler camera and performs iterative denoising to predict a sequence of 16 actions at a time. The predicted actions and the corresponding observation timestamps are sent to the rollout process. At every step of policy execution, the rollout process receives an action sequence from the inference process and then skips a few actions so that the predicted actions sent to the robot are aligned in time with what the policy expects the robot to execute. During evaluation, we use the same DDPM noise scheduler as the training procedure.

\textbf{Details of the \textit{Humans} baseline.}
For each task, we ask 10 unique users to drive the mobile robot to where they believe is the optimal starting base pose to execute the task. All users have technical backgrounds in robotics, but are not aware of how the policies are trained and what robot positions are best for policy execution. We then initialize the robot at each of these human-designated navigation targets, and then evaluate the manipulation policy from there. This allows us to test whether our method can even outperform navigation targets generated via human intuition.

\textbf{Measuring task success.} For each task and method, we track the incidence rates of certain performance milestones (i.e., navigation success, grasp success, full success), and also of task-ending collisions (i.e., fatal failures). We consider a navigation attempt successful if the robot drives to a pose that not only faces the target object but also is within a distance of 50~cm from the ideal target pose. A grasp success is defined as when the robot stably grasps the object of interest (i.e. the chip bag, the paper towels, or the tray of tomatoes), and full success is counted only if the entire task is completed.

\textbf{Detailed analysis of results.} We find that for all three manipulation tasks, our method significantly outperforms the two baselines. The end-to-end \textit{BC w/ Nav} baseline performs decently well in the \textit{Pour Tomatoes} task, where the robot can navigate to a reasonable position $80\%$ of the time, ultimately achieving a full success rate of $70\%$ with only $20\%$ of the runs ending in collision. In the \textit{Bimanual Grab Paper Towels} and \textit{Pick Chips} tasks, however, it sees an uptick in fatal failures at $70\%$ and $40\%$ respectively. In our experiments, \textit{BC w/ Nav} only achieves a $30\%$ navigation success rate for both of these two tasks, which even further degrades into a $0\%$ full success rate for the \textit{Pick Chips} task.

The human baseline achieves a very high navigation success of at least $80\%$ for all tasks. However, the success rate of the entire task is very low, with a maximum score of $40\%$ in \textit{Bimanual Grab Paper Towels}. In the \textit{Pick Chips} task, $20\%$ of the rollouts even end in arm-collisions with the shelf. This result yields an interesting insight---although humans are adept at intuitively navigating the robot into very reasonable starting poses, they are ultimately unaware of what the policy training data looks like. Therefore, the human-provided navigation poses still generally fail to achieve task success. Our method is able to pick up on the subtleties of these distribution shifts. In real experiments, our method achieves successful final navigation poses $100\%$ of the time for all three tasks, and achieves full success rates of $70\%$, $100\%$, and $80\%$ for the \textit{Pick Chips}, \textit{Bimanual Grab Paper Towels}, and \textit{Pour Tomatoes} tasks, respectively. Furthermore, unlike any of the evaluated baselines, the rollouts using our method are completely collision-free. 

\section{Additional Visualizations}

\begin{figure*}[t]
    \centering
    \begin{subfigure}[t]{0.32\textwidth}
        \includegraphics[width=\linewidth]{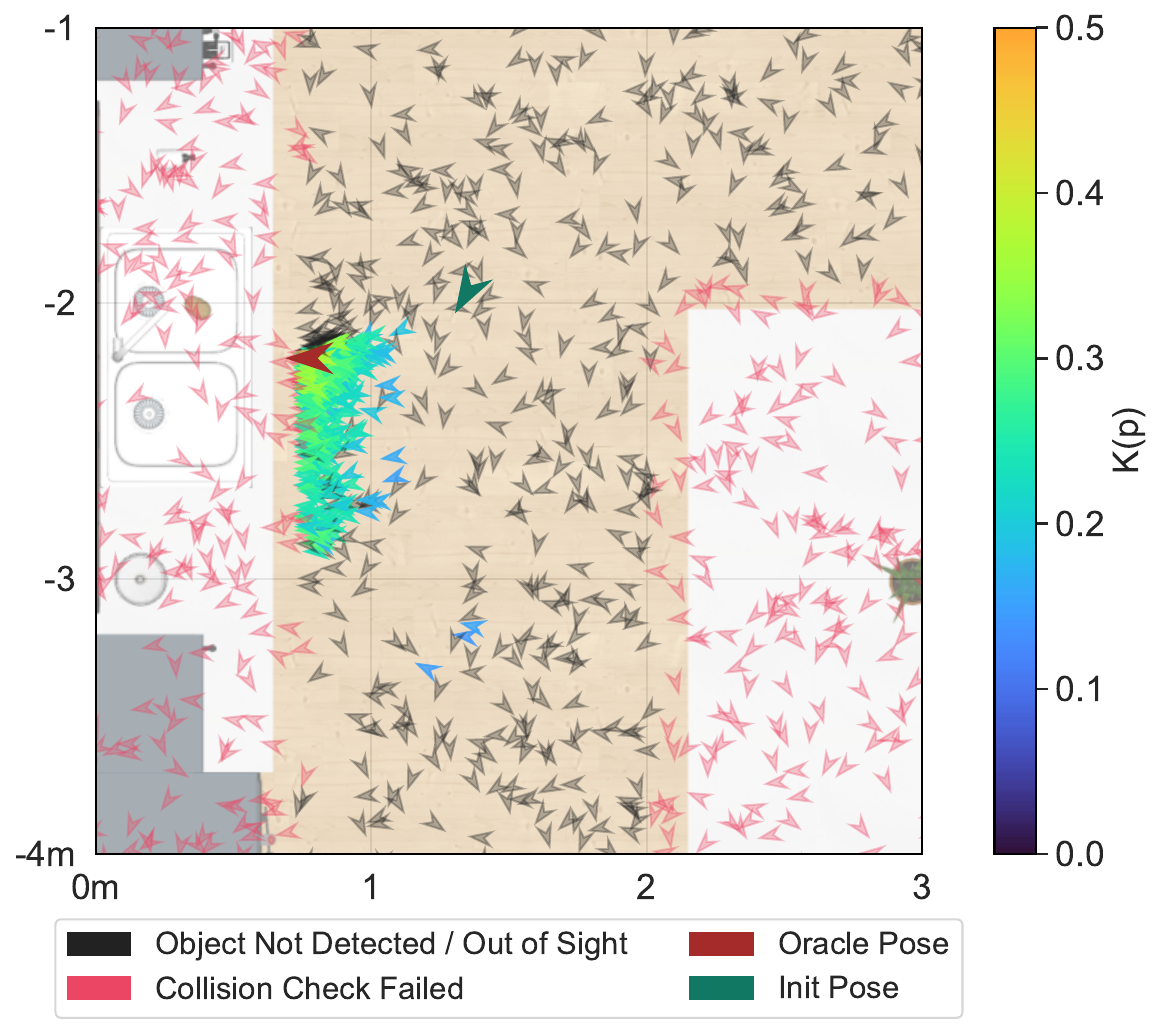}
    \end{subfigure}
    \begin{subfigure}[t]{0.32\textwidth}
        \includegraphics[width=\linewidth]{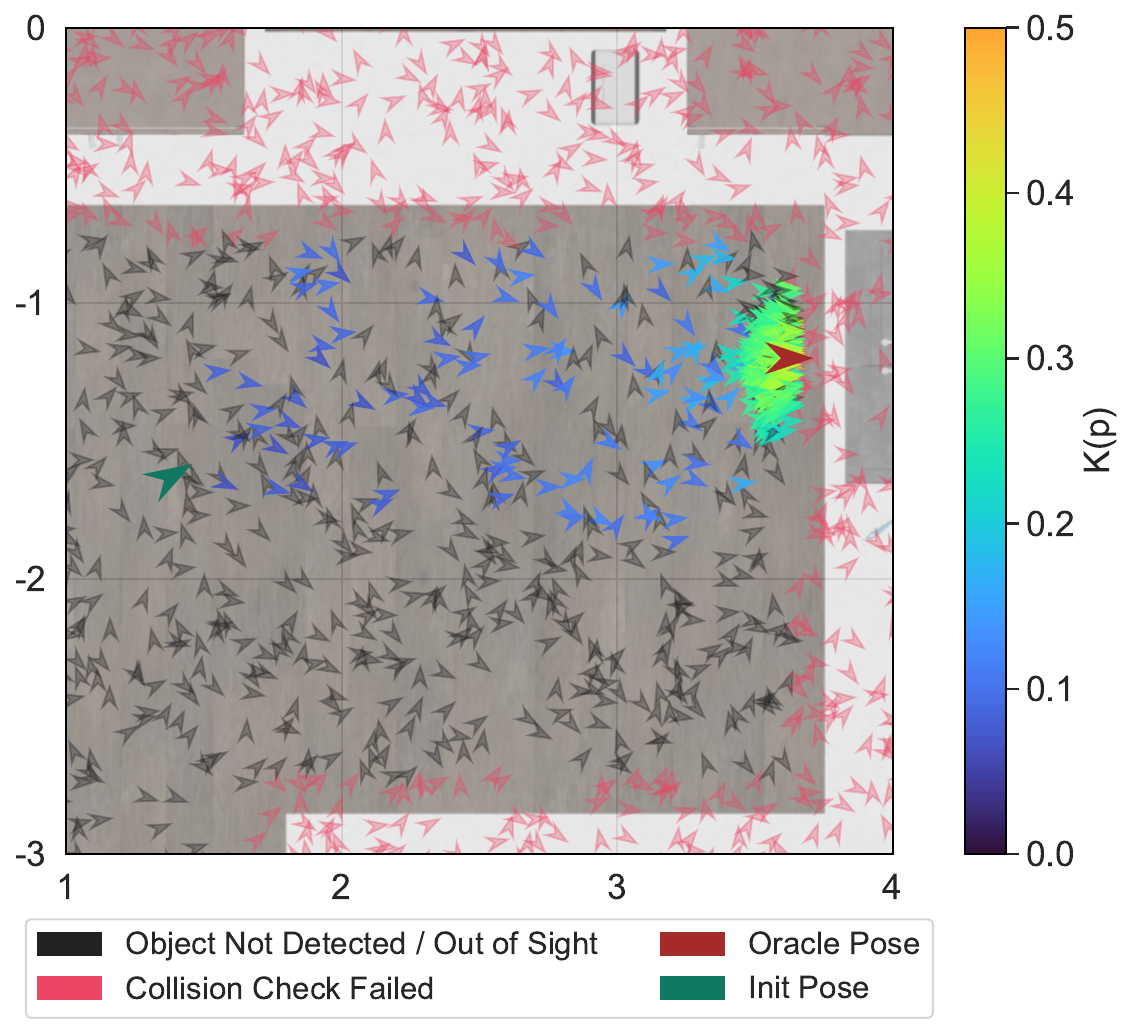}
    \end{subfigure}
    \begin{subfigure}[t]{0.32\textwidth}
        \includegraphics[width=\linewidth]{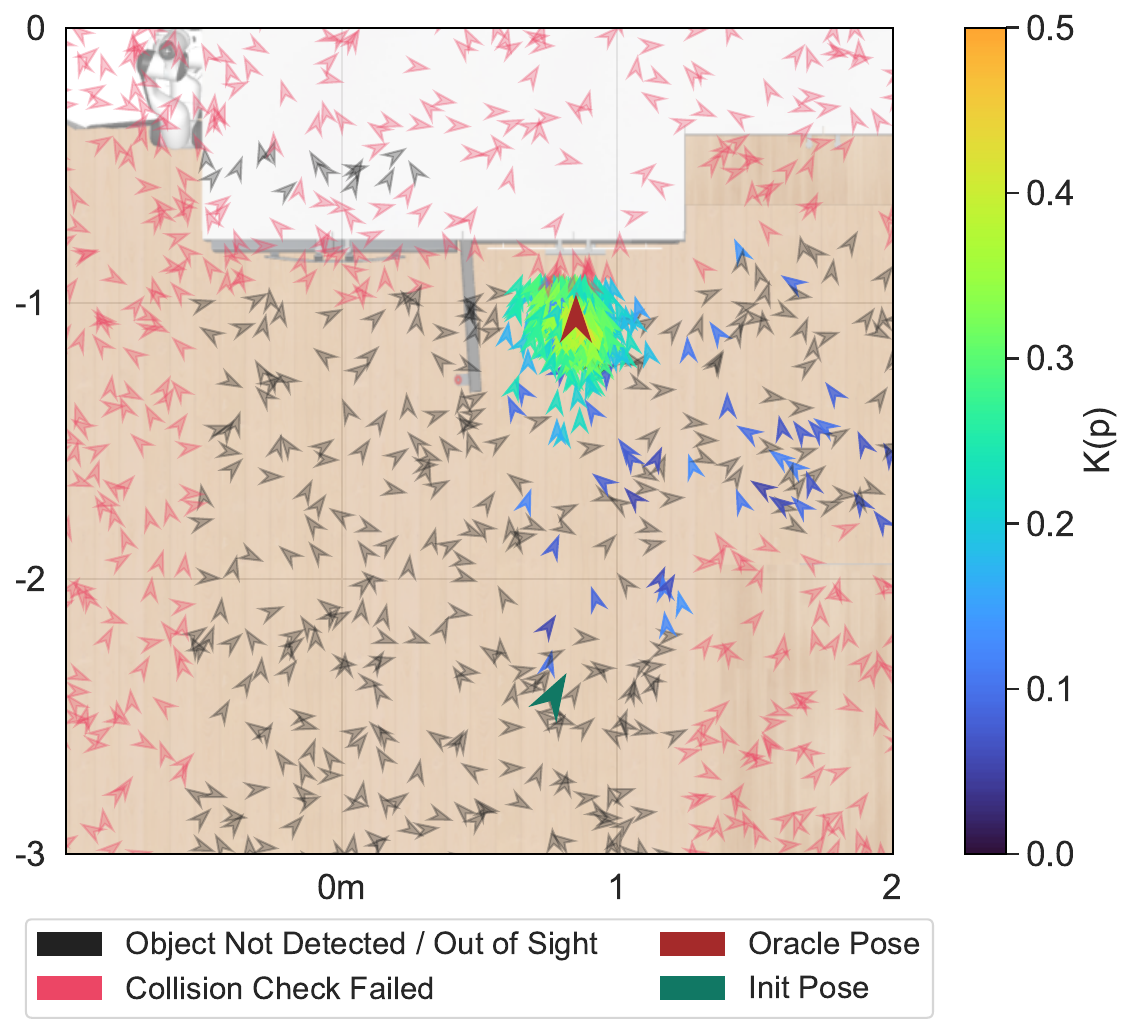}
    \end{subfigure}
    \caption{\textbf{Visualizations of policy mobilization process of our method.} Each plot shows one episode of robot pose optimization for policy mobilization. The background plots the top-down map of the environment where the robot needs to run the designated policy. Each arrow represents the position and heading of a robot pose sampled by our method. In the plot, our method converges toward the oracle policy initialization robot pose (in dark red).}
    \label{fig:score_maps_more}
    \vspace{-10pt}
\end{figure*}

In addition to the visualizations shown in the main paper, we also design tools to visualize the pose optimization process. In Figure~\ref{fig:score_maps_more}, we show top-down maps of the simulated benchmark environment overlaid with arrows and markers illustrating the policy mobilization process of our method. 

First, we plot the oracle pose for executing the policy in dark red and the initial pose in dark green. Then, we plot the position and heading of robot poses sampled by our method, along with the predicted scores represented by color codes. Areas with collision check failures are marked pink; areas where the target object is not in view are marked black, and areas where $K_\text{id}$ predicts a score are marked with colors in a rainbow color map that represent the predicted scores. The figure shows an example of our method sampling throughout the scene and converging to the optimal robot pose for starting the execution of the manipulation policy. Such visualizations make debugging and developing policy mobilization methods easier with clear and precise visual feedback.

\end{document}